\pdfoutput=1
\documentclass[11pt,a4paper]{article}
\usepackage[hyperref]{acl2021}

\usepackage{times}
\usepackage{latexsym}
\usepackage{cancel}
\usepackage{graphicx}
\usepackage{amsmath}
\usepackage{latexsym}
\usepackage{amssymb}
\usepackage{amsthm}
\usepackage{color}
\usepackage{xspace}
\usepackage{subfigure}
\usepackage{array, booktabs, makecell}

\usepackage{booktabs}
\usepackage{array,tabularx}
\usepackage[ruled,vlined]{algorithm2e}
\usepackage{algpseudocode}

\usepackage{ulem}
\usepackage{amssymb}% http://ctan.org/pkg/amssymb
\usepackage{pifont}% http://ctan.org/pkg/pifont
\usepackage{multirow}

\newenvironment{des}{                 % list without par spacings
     \parskip 0cm \begin{list}{}{\parsep 0cm \itemsep 0cm \topsep 0cm}}{
       \end{list}} %  \parskip 0cm}

\newenvironment{conditions*}
  {\par\vspace{\abovedisplayskip}\noindent
   \tabularx{\columnwidth}{>{$}l<{$} @{}>{${}}c<{{}$}@{} >{\raggedright\arraybackslash}X}}
  {\endtabularx\par\vspace{\belowdisplayskip}}

\usepackage{microtype}

\aclfinalcopy % Uncomment this line for the final submission
\title{CURIE: An Iterative Querying Approach for Reasoning About Situations }

\author{Dheeraj Rajagopal~\thanks{\hspace{0.5em} authors contributed equally to this work. Ordering determined by dice rolling.}\hspace{0.5em}, Aman Madaan~\footnotemark[1]\hspace{0.5em}, Niket Tandon$^\dagger$,  Yiming Yang,\\ 
\textbf{Shrimai Prabhumoye, Abhilasha Ravichander, Peter Clark$^\dagger$}, \textbf{Eduard Hovy} \\
  Language Technologies Institute, Carnegie Mellon University \\
  Pittsburgh, PA, USA \\ 
  $^\dagger$ Allen Institute for Artificial Intelligence \\
  Seattle, WA, USA \\ 
  \texttt{\{dheeraj,amadaan,yiming,sprabhum,aravicha,hovy\}@cs.cmu.edu} \\ \texttt{\{nikett, peterc\}@allenai.org} \\}

\begin{document}
\definecolor{Red}{rgb}{1,0,0}
\definecolor{Green}{rgb}{0,1,0}
\definecolor{Blue}{rgb}{0,0,1}
\definecolor{Red}{rgb}{0.9,0,0}
\definecolor{Orange}{rgb}{1,0.5,0}
\definecolor{yellow}{rgb}{0.65,0.6,0}
\definecolor{cadmiumgreen}{rgb}{0.2, 0.7, 0.24}

\newcommand{\nk}[1]{\textcolor{Red}{[#1 \textsc{--Niket}]}}
\newcommand{\pc}[1]{\textcolor{Red}{[#1 \textsc{--Peter}]}}
\newcommand{\lasha}[1]{\textcolor{yellow}{[#1 \textsc{--Abhilasha}]}}
\newcommand{\am}[1]{\textcolor{Orange}{[#1 \textsc{--Aman}]}}
\newcommand{\dheeraj}[1]{\textcolor{Blue}{[#1 \textsc{--Dheeraj}]}}
\newcommand{\sm}[1]{\textcolor{Orange}{[#1 \textsc{--SM}]}}
\newcommand{\notes}[1]{\textcolor{Red}{[#1 \textsc{--notes}]}}
\newcommand{\ed}[1]{\textcolor{cyan}{[#1 \textsc{--Ed}]}}

\newcommand{\ar}[1]{\textcolor{red}{\bf\small [#1 --Abhilasha]}}
\newcommand{\secref}[1]{\S\ref{#1}}
\newcommand\given[1][]{\:#1\vert\:}

\newcommand{\V}[1]{\mathbf{#1}}

%Constants
\newcommand{\numgraphs}{2107\xspace}
\newcommand{\numpassages}{379\xspace}
\newcommand{\numquestions}{40.7K\xspace}
\newcommand{\grn}[1]{\textcolor{cadmiumgreen}{#1}}
\newcommand{\red}[1]{\textcolor{Red}{#1}}

%Hyperparams
\newcommand{\lrate}{\textcolor{Red}{LR-HERE} }
\newcommand{\dropout}{\textcolor{Red}{DROPOUT-HERE} }
\newcommand{\green}[1]{\textcolor{green}{#1}}
\newcommand{\cadmiumgreen}[1]{\textcolor{cadmiumgreen}{#1}}

\newcommand{\helps}{\overset{+}{\longrightarrow}}
\newcommand{\hurts}{\overset{-}{\longrightarrow}}
\newcommand{\helpedby}{\overset{+}{\longleftarrow}}
\newcommand{\hurtby}{\overset{-}{\longleftarrow}}
\newcommand{\entails}{\overset{\oplus}{\longrightarrow}}
\newcommand{\doesnotentail}{\overset{\oplus}{\longleftarrow}}
\newcommand{\relatedby}{\overset{r}{\longrightarrow}}
\newcommand{\ir}{\textsc{ir}\xspace}
\newcommand{\wiqa}{\textsc{wiqa}\xspace}
\newcommand{\quartz}{\textsc{quartz}\xspace}
\newcommand{\quarel}{\textsc{quarel}\xspace}
\newcommand{\defeasible}{\textsc{defeasible}\xspace}
\newcommand{\bert}{\textsc{wiqa-bert}\xspace}
\newcommand{\qa}{\textsc{qa}\xspace}
\newcommand{\ours}{\textsc{curie}\xspace}
\newcommand{\comet}{\textsc{comet}\xspace}
\newcommand{\plm}{\textsc{plm}\xspace}
\newcommand{\cmark}{\ding{51}}%
\newcommand{\xmark}{\ding{55}}%
\newcommand{\lms}{\textsc{lm}\xspace}
\newcommand{\rqone}{\textsc{rq1}\xspace}
\newcommand{\rqtwo}{\textsc{rq2}\xspace}

\newcommand{\gpt}{\texttt{GPT}\xspace}
\newcommand{\gptt}{\texttt{GPT-2}\xspace}
\newcommand{\cf}{\textit{st}\xspace}
\newcommand{\st}{\textit{st}\xspace}
\newcommand{\bleu}{\texttt{BLEU}\xspace}
\newcommand{\rouge}{\texttt{ROUGE}\xspace}
\newcommand{\igen}{\textsc{IterativeGraphGen}\xspace}
\newcommand{\igencall}[3]{\textsc{IterativeGraphGen}(#1, #2, #3)\xspace}
\newcommand{\lm}{$\mathcal{M}$\xspace}

\newcommand{\squishlist}{
  \begin{list}{$\bullet$}
    { \setlength{\itemsep}{0pt}      \setlength{\parsep}{3pt}
      \setlength{\topsep}{3pt}       \setlength{\partopsep}{0pt}
      \setlength{\leftmargin}{1.5em} \setlength{\labelwidth}{1em}
      \setlength{\labelsep}{0.5em} } }
\newcommand{\reallysquishlist}{
  \begin{list}{$\bullet$}
    { \setlength{\itemsep}{0pt}    \setlength{\parsep}{0pt}
      \setlength{\topsep}{0pt}     \setlength{\partopsep}{0pt}
      \setlength{\leftmargin}{0.2em} \setlength{\labelwidth}{0.2em}
      \setlength{\labelsep}{0.2em} } }

 \newcommand{\squishend}{
     \end{list} 
 }

\maketitle
\begin{abstract}
Recently, models have been shown to predict the effects of unexpected situations, e.g., would cloudy skies help or hinder plant growth? Given a context, the goal of such situational reasoning is to elicit the consequences of a new situation (\st) that arises in that context. We propose a method to iteratively build a graph of relevant consequences explicitly in a structured situational graph (\st graph) using natural language queries over a finetuned language model (\lm). Across multiple domains, \ours generates \st graphs that humans find relevant and meaningful in eliciting the consequences of a new situation. We show that \st graphs generated by \ours improve a situational reasoning end task (\wiqa-\qa) by 3 points on accuracy by simply augmenting their input with our generated situational graphs, especially for a hard subset that requires background knowledge and multi-hop reasoning.

\end{abstract}

\normalem

\section{Introduction}

A long-standing challenge in reasoning is to model the consequences of a novel situation in a context.
Consider these questions - \emph{Would it rain more if we plant more trees?}, or  \emph{What would help water to boil faster?} - answering these questions requires comprehending the complex events such as \emph{plant growth} and \emph{water boiling}, where much of the information remains implicit (by Grice's maxim of quantity \citep{Grice1975LogicAC}), thus requiring inference.

\begin{figure}[t]
{\includegraphics[width=1.1\columnwidth]{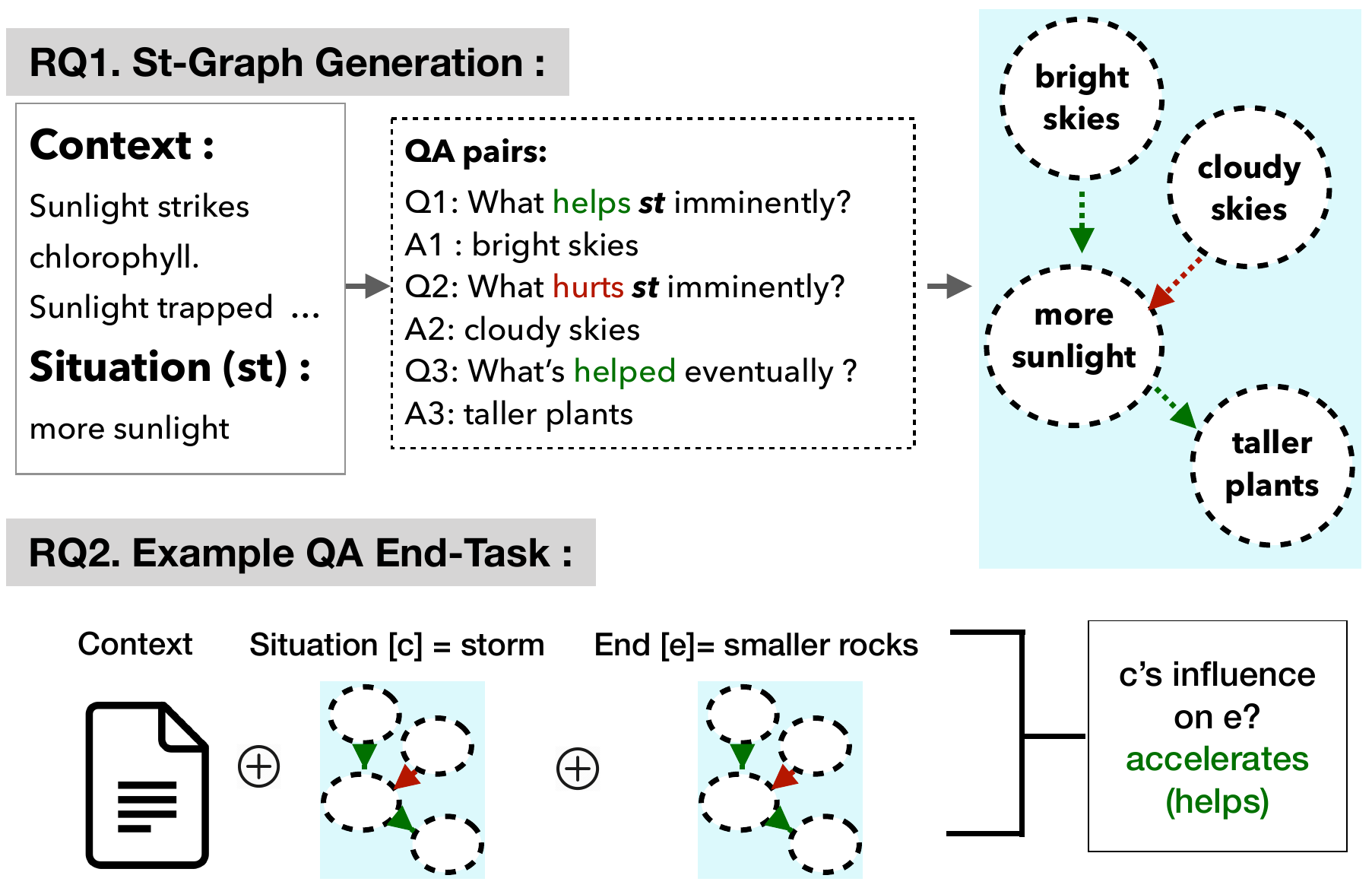}}
\caption{
\rqone: \ours generates situational graphs through iterative queries to a model, making the model’s knowledge of influences explicit (above;  \grn{positive}, and \red{negative} influence) iteratively. \rqtwo: Such graphs can improve situational reasoning QA when added to the QA input (below, where the context is a passage about erosion).
}
\label{fig:intro-example}
\end{figure}

Tasks that require situational reasoning are increasingly observed by machines deployed in the real world - unexpected situations are common, and machines are expected to gracefully handle them. It is also essential for tasks such as qualitative reasoning ~\citep{tandon2019wiqa,tafjord2019quarel}, physical commonsense reasoning tasks~\citep{sap2019atomic,bisk2020piqa}, and defeasible inference~\citep{rudinger-etal-2020-thinking}.
Unlike humans, machines are not adept at such reasoning.

Prior systems that address situational reasoning take as input a context providing background information, a situation (\cf),  \textit{and} an ending, and predict the reachability from \cf to that ending either in a classification setting (e.g., \citet{tandon2019wiqa} grounds the path on at most two sentences in the context) or recently, in a story-generation setting~\cite{counterfactual-story-generation-2019-emnlp}, where the goal is to generate an alternate ending when the original ending and a counterfactual situation are given.
However, generating effects of situations in real-world scenarios, where the ending is typically unknown is still an open challenge.
We also might need \cf-reasoning capabilities across multiple domains (beyond stories).
Further, multiple types of consequences to a situation might have to be generated (e.g,  positive and negative impacts or eventual and immediate impacts), which requires outputs in a structured form.

To address these limitations, we propose \ours - a generation framework that generalizes multiple reasoning tasks under a general situational reasoning framework. 
The task is illustrated in Figure \ref{fig:intro-example}: given some context and just a situation \cf (short phrase), our framework generates a \textit{situational reasoning graph} (\cf-graph). At its core, \ours constructs a reasoning graph based on the contextual knowledge that supports the following kinds of reasoning:
 \begin{des}
     \item[1.] If \cf occurs, what will happen imminently/ eventually?
     \item[2.] If \cf occurs, which imminent/ eventual effect will not happen?
     \item[3.] What will support/ prevent the \cf?
 \end{des}

As shown in Figure \ref{fig:intro-example}, our approach to this task is to iteratively compile the answers to questions 1,2,3 to construct the \cf-graph. Compared to a free-form text output obtained from an out-of-the-box seq-to-seq model, our approach gives more control and flexibility over the graph generation process, including arbitrarily reasoning for any particular node in the graph. Downstream tasks that require reasoning about situations can compose natural language queries to construct a \cf-reasoning graph that can be simply augmented to their input.
In this paper, we ask the following two research questions:

\squishlist
\item[\textbf{\rqone}] Given a specific context and situation, can we iteratively generate a situational reasoning graph of potential effects?
\item[\textbf{\rqtwo}] Can the \st-graphs generated by \ours improve performance at a downstream task? 
\squishend
In response, we make the following contributions:
\begin{des}
\item[(i.)] We present \ours, the first domain-agnostic situational reasoning framework that takes as input some context and an \st and iteratively generates a situational reasoning graph ~(\S{\ref{sec:approach}}). We show that our framework is effective at situational reasoning across three datasets, as validated by human evaluation and automated metrics.

\item[(ii.)] We show that \st graphs generated by \ours improve a \st-reasoning task (\wiqa-\qa) by 3 points on accuracy by simply augmenting their input with our generated situational graphs, especially for a hard subset that requires background knowledge and multi-hop reasoning~(\S{\ref{sec:downstreamqa}}). (Table \ref{tab:research_questions}).
%\footnote{anonymized code and data are located at:\\ \url{https://tinyurl.com/curie-acl}}

\end{des}

\begin{table*}[!ht]
\small
\centering
\resizebox{\textwidth}{!}{%
\begin{tabular}[b]{@{}llll@{}} % top alignment
\toprule
Dataset & Original formulation & \cf formulation & \cf graph     
 \\ \midrule

\small{\wiqa} & \begin{tabular}[b]{@{}l@{}}\textit{context}: Wind creates waves.. \\ Waves wash on beaches... \\ \textit{ques}: If there is storm, how \\  will it affect bigger waves?\\ \textit{chain}: storm $\rightarrow$ stronger \\ wind $\rightarrow$ bigger waves\\ \textit{answer}: bigger waves\end{tabular} &  \begin{tabular}[b]{@{}l@{}}Given \textit{context} and \\ \cf: there is a storm\\ Q1: What does \cf \cadmiumgreen{help} \textit{imminently} ? \\ A1: stronger wind\\ Q2: What does \cf \cadmiumgreen{help} \textit{eventually} ?\\ A2: bigger waves\end{tabular} & \includegraphics[scale=0.4, height=25mm]{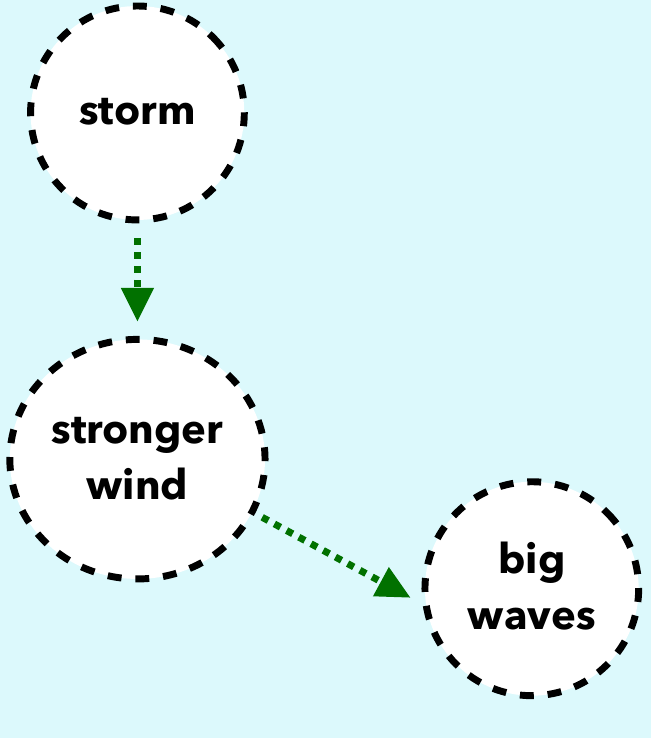} \\ \midrule

\small{\quarel} & \begin{tabular}[b]{@{}l@{}}\textit{context}: Car rolls further on \\  wood than on thick carpet\\ \textit{ques}: what has more resistance? \\ (a) wood (b) the carpet \\ \textit{simplified logical form of} \\ \textit{context, ques}: \\ distance is higher on wood $\rightarrow$ \\ (a) friction is higher in carpet (or)\\ (b) friction is higher in wood\\ \textit{answer}: (b) the carpet \end{tabular} & \begin{tabular}[b]{@{}l@{}}Given \textit{context} and\\ \cf : distance is higher on wood\\ Q1: What does \cf  \cadmiumgreen{entail} \textit{imminently} ? \\ A1: friction is lower in wood\\ Q2: What does \cf \red{contradict} \textit{imminently} ? \\ A2: friction is lower in carpet\\ Q3: What does \cf \cadmiumgreen{entail} \textit{eventually} ? \\ A3: wood has more resistance\end{tabular}  & \includegraphics[scale=0.4, height=25mm]{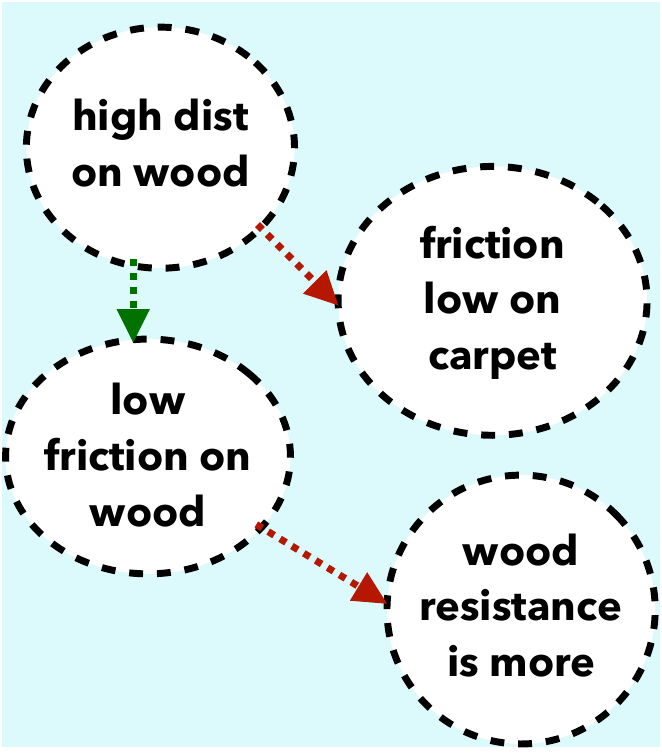} \\ \midrule

\small{\textsc{defeas}} & \begin{tabular}[b]{@{}l@{}}\textit{context}: Two men and a dog are  \\             standing  among the green hills.\\ \textit{hypothesis}: The men are farmers.\\ \textit{evidence type}: strengthener\\ \textit{answer}: the dog is a sheep dog\end{tabular} & \begin{tabular}[b]{@{}l@{}}Given \textit{context} and \\ \cf : dog is a sheep dog\\ Q1: What does \cf \cadmiumgreen{strengthen} \textit{imminently} ? \\ A1: The men are farmers\\ \cf : men are studying tour maps\\ Q2: What does \cf \red{weaken} \textit{imminently}? \\ A2: The men are farmers\end{tabular} & \includegraphics[scale=0.4, height=25mm]{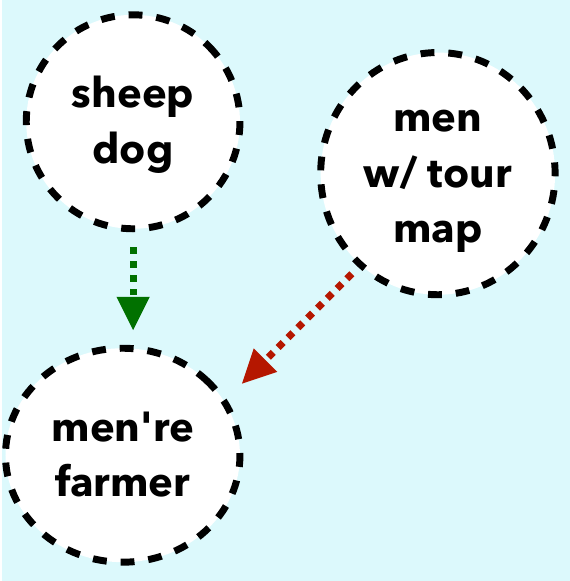} \\ \bottomrule
\end{tabular}%
}
\caption{The datasets used by \ours and how we re-purpose them for \cf reasoning graph generation task. As explained in \S{\ref{subsec:task}}, the \grn{green} edges set depicts relation ($r$) (entail, strengthen, helps) and \red{red} edges depict one of (contradict, weaken, hurts). The \{ \textit{imminent}, \textit{eventual} \} effects ($c$) are used to support multihop reasoning. \textsc{defeas} = \defeasible, \emph{chain} refers to reasoning chain. Some examples are cut to fit.
The key insight is that an \st-graph can be decomposed into a series of QA pairs, enabling us to leverage seq-to-seq approaches for \st-reasoning.
}
\label{tab:dataset_creation}
\end{table*}

\section{\ours for Situational Reasoning}
\label{sec:approach}

\begin{figure}[tb]
{\includegraphics[width=1.0\columnwidth]{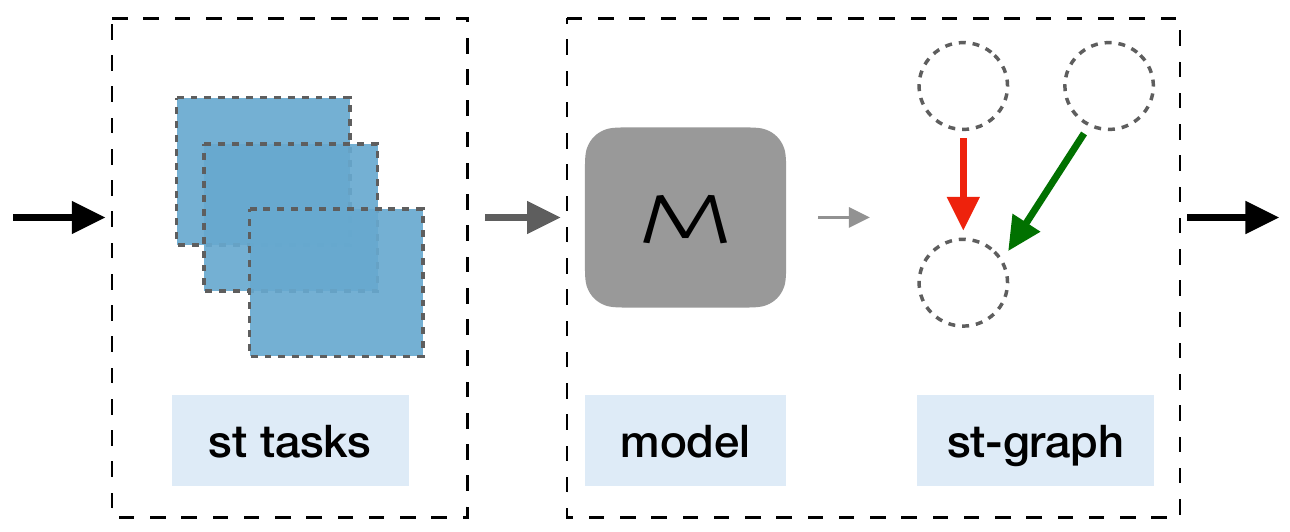}}
\caption{\ours framework consists of two components: (i) a formulation that adapts datasets that allow \cf-reasoning for pretraining (ii) a method to iteratively build structured \st-graphs using natural language queries over a fine-tuned language model (\lm).}
\label{fig:curie_architecture}
\end{figure}

\ours provides both a general framework for situational reasoning and a method for constructing \cf-reasoning graphs from pretrained language models.
The overall architecture of \ours is shown in Figure \ref{fig:curie_architecture}. \ours framework consists of two components: (i) \emph{\st-reasoning task formulation :} a formulation that adapts datasets that allow situational reasoning  (ii) \emph{\st-graph construction :} a method to fine-tune language  model \lm to generate the consequences of a situation and iteratively construct structured situational graphs (shown in figure~\ref{tab:dataset_creation}). 
In this section, we present (i) our task formulation (\secref{subsec:task}), (ii) adapting existing datasets for \ours task formulation~(\secref{subsec:dataset}), (iii) the learning procedure (\secref{subsec:masked-lm}), and (iv) the \cf-graph generation via inference (\secref{subsec:inference}).

\subsection{Task Formulation}
\label{subsec:task}

We describe the general task formulation for adapting pretraining language models to the \st-reasoning task.  
Given a context $T = \{ s_1, s_2,\dots,s_N \}$ with $N$ sentences, and a situation \st, our goal is to generate an \st-graph $G$ in this changed world.

An \st-graph $G(V, E)$ is an unweighted directed acyclic graph. 
A vertex $v \in V$ is an event or a state such that it describes a change to the original conditions in $T$. 
Each edge $e_{ij} \in E$ is labeled with an relationship $r_{ij}$, that indicates whether $v_i$ \emph{positively} or \emph{negatively} influences $v_j$. 
Positive influences are represented via \grn{green} edges comprising one of \{\textit{entails, strengthens, helps}\} and negative influences represented via \textcolor{Red}{red} edges that depict one of \{\textit{contradicts, weakens, hurts}\}.
Our relation set is general and can accommodate various \cf-reasoning tasks. 
Given two nodes $v_i, v_k \in V$, if a path from $v_i$ to $v_k$ has more than one edge, we describe the effect $c$ as \emph{eventual} and a direct effect as \emph{imminent}.

We obtain the training data for st-graph generation task by decomposing an \st-graph into a set of question-answer pairs.
Each question comprises of the context $T$, a \cf-vertex $v_s$, a relation $r$, and the nature of the effect $c$. 
The output is an answer to the question, that corresponds to the target node $v_t$.
An example is shown in Figure \ref{fig:intro-example}. 
Compared to an end-to-end approach to graph generation, our approach gives more flexibility over the generation process, enabling reasoning for any chosen node in the graph.

\subsection{Generalizing Existing Datasets}
\label{subsec:dataset}

Despite theoretical advances, lack of large-scale general situational reasoning datasets presents a challenge to train seq-to-seq language models. 
In this section, we describe how we generalize existing diverse datasets towards \cf-reasoning towards finetuning a language model $\mathcal{M}$. 
If a reasoning task allows a context, a \cf-situation and can describe the influence of \cf in terms of \grn{green} and/or \red{red} edges, it can be seamlessly adapted to \ours framework.
Due to lack of existing datasets that directly support our task formulation, 
adapt the following three diverse datasets - \wiqa, \quarel and \defeasible for \ours.

\paragraph{\wiqa :} \wiqa task studies the effect of a perturbation in a procedural text \citep{tandon2019wiqa}.
The context $T$ in \wiqa is a procedural text describing a physical process, and \cf is a perturbation i.e., an external situation deviating from $T$, and the effect of \cf is either \grn{helps} or \red{hurts}. An example of WIQA to \cf-formulation is shown in Table \ref{tab:dataset_creation}.

\begin{table*}[!h]
\centering
{
    \begin{tabular}{l l l l l}
    \toprule
    \textbf{Research question} & \textbf{Training dataset} & \textbf{Test dataset} & \textbf{Task} & \textbf{Metrics} \\
        \toprule
        Can we generate & \wiqa-\cf        & \wiqa-\cf           & generation & \rouge, \bleu \\
        good \cf graphs?    (\S{\ref{sec:experiments}})    & \quarel-\cf      & \quarel-\cf         & generation & \rouge, \bleu \\
                                          & \defeasible-\cf  & \defeasible-\cf     & generation & \rouge, \bleu \\
        \midrule
        Can we improve  & \wiqa-\cf, \wiqa-QA  & \wiqa-QA        & finetuned QA & accuracy \\
        downstream tasks?   (\S{\ref{subsec:wiqa-qa}}, \secref{sec:zeroshot})  & \quarel-\cf  & \quartz-QA       & zero shot & accuracy \\
        \bottomrule
    \end{tabular}
}
\caption{Overview of experiments}
    \centering
    \label{tab:research_questions}
\end{table*}

\begin{table}[!h]
\centering
\begin{tabular}{llll}
\toprule
Dataset & train &  dev & test\\ \midrule
\wiqa & 119.2k & 34.8k  & 34.8k \\ 
\quarel &  4.6k & 1.3k & 652 \\ 
\defeasible & 200k  & 14.9k & 15.4k\\ 
\bottomrule
\end{tabular}
\caption{Dataset wise statistics, we maintain the splits}
\label{table:data-split}
\end{table}

\paragraph{\quarel :} \quarel dataset \cite{tafjord2019quarel} contains qualitative story questions where $T$ is a narrative, and the \cf is a qualitative statement. $T$ and \cf are also expressed in a simpler, logical form, which we make use of because it clearly highlights the reasoning challenge. The effect of \cf is either \grn{entails} or \red{contradicts} (example in Table \ref{tab:dataset_creation}).

\paragraph{\defeasible :}
The \defeasible reasoning task \citep{rudinger-etal-2020-thinking} studies inference in the presence of a counterfactual. The context $T$ is given by a premise which describes a everyday context, and the \cf is an observed evidence which either \grn{strengthens} or \red{weakens} the hypothesis. We adapt the original abductive setup as shown in Table \ref{fig:intro-example}. 
In addition to commonsense situations, \defeasible-\cf also comprises of social situations, thereby contributing to the diversity of our datasets.

\subsection{Learning to Generate \cf-graphs}
\label{subsec:masked-lm}

To reiterate our task formulation (\secref{subsec:task}), for a given context and \cf, we first specify 
a set of questions and the resulting output for the questions is then compiled to form a \cf-graph.

The training data thus consists of tuples $(\mathbf{x}_i, \mathbf{y}_i)$, with $\mathbf{x}_i = (T, \cf~, r, c)_i$ where $T$ denotes the context, \cf the situation, $r$ denotes the edge (\grn{green} or \red{red}), $c$ signifies the nature of the effect (imminent or eventual), and $\mathbf{y}_i$ is the output (a short sentence or a phrase depicting the effect).
The output of $N_Q$ such questions is compiled into a graph $G = \{ \V{y_i} \}_{1:N_Q}$ (as shown in Figure \ref{fig:intro-example}).

We use a pretrained language model $\mathcal{M}$ to estimate the probability of generating an answer $\mathbf{y}_i$ for an input $\mathbf{x}_i$. We first transform the tuple $\mathbf{x}_i = \langle x_i^{1}, x_i^{2},\ldots, x_i^{N} \rangle$ into a single query sequence of tokens by concatenating its components i.e. we set $\mathbf{x}_i = \texttt{concat} (T, \cf~,r,c) $, where \texttt{concat} refers to string concatenation.
Let the sequence of tokens representing the target event be $\mathbf{y}_i = \langle y_i^{1}, y_i^{2},\ldots, y_i^{M} \rangle$, where $N$ and $M$ are the lengths of the query and the target event sequences %(\red{N is not defined --NT xxxxxxxxxxxxxx}).
We model the conditional probability $p_{\theta}(\mathbf{y}_i \given \mathbf{x}_i)$ as a series of conditional next token distributions parameterized by $\theta$: as $ p_{\theta}(\mathbf{y}_i \given \mathbf{x}_i) =  \prod_{k=1}^{M} p_{\theta} (y_i^k \given \mathbf{x}_i, y_i^{1},.., y_i^{k-1}) $.

\SetKwInput{KwGiven}{Given}
\SetKwInput{KwInit}{Init}
\begin{algorithm}[tb]
\SetAlgoLined
\KwGiven{\ours language model $\mathcal{M}$.}
\KwGiven{Context passage $T$, a situation \cf, a set $R = \{(r_i, c_i)\}_{i=1}^{N_Q}$ made of $N_Q$ $(r, c)$ tuples.}
\KwResult{\cf graph $G$ where the $i^{th}$ node will be generated with the relation $r_i$ and the effect type $c_i$.}

\KwInit{$G \gets \emptyset$}
\For{$i \gets 1, 2, \ldots, N_Q$}{
\tcp{Create a query}
$\mathbf{x}_i = \texttt{concat} (T, \cf,r_i,c_i) $\;
\tcc{Sample a node from the language model $\mathcal{M}$}
$\mathbf{y}_i \sim \mathcal{M}(\mathbf{x}_i)$\;
\tcc{Add the sampled node and the edge to the graph}
$G = G \cup (r_i, c_i, \mathbf{y}_i$)\;
}
\KwRet{$G$}

\caption{\igen(\texttt{IGEN}): generating \cf graphs with \ours}
\label{alg:alg-1}
\end{algorithm}

\subsection{Inference to Decode \st-graphs}
\label{subsec:inference}

The auto-regressive factorization of the language model $p_{\theta}$ allows us to efficiently generate target event influences for a given test input $\mathbf{x}_j$.

The process of decoding begins by sampling the first token $y_j^{1} \sim p_\theta(y \given \mathbf{x}_j)$.
The next token is then drawn by sampling $y_j^{2} \sim p_\theta(y  \given \mathbf{x}_j, y_j^{1})$.
The process is repeated until a specified \emph{end-symbol} token is drawn at the $K^{th}$ step.
We use nucleus sampling~\citep{holtzman2019curious} in practice.
The tokens $\langle y_j^{1}, y_j^{2},\ldots, y_j^{K - 1}\rangle$ are then returned as the generated answer.
To generate the final \cf-reasoning graph $G$, we combine all the generated answers $\{ \mathbf{y}_i \}_{1:N_Q}$ that had the same context and \cf pair $(T, \cf~)$ over all $(r,c)$ combinations.
We can then use generated answer $\cf~' \in \{ \mathbf{y}_i \}_{1:N_Q}$, as a new input to $\mathcal{M}$ as $(T, \cf~')$ to recursively expand the \cf-graph to arbitrary depth and structures~(Algorithm~\ref{alg:alg-1}). One such instance of using \ours \cf graphs for a downstream QA task is shown in \secref{sec:downstreamqa}.

\section{\rqone: Establishing Baselines for \st-graph Generation}
\label{sec:experiments}

This section reports on the quality of the generated \cf reasoning graphs and establishes strong baseline scores for \st-graph generation.

We use the datasets described in section \secref{subsec:dataset} for our experiments. 

\begin{table}[]
\centering
% \small
\begin{tabular}{lrr}
\hline
Model (\lm)                                        & BLEU                          & ROUGE                           \\ \hline
\multicolumn{1}{c}{\wiqa-\cf}       &                                 &                                 \\ \hline
\textsc{lstm} Seq-to-Seq                         & 7.51                            & 18.71                           \\
\gpt$\sim$(w/o $T$)                                & 7.82                            & 19.30                           \\
\gptt$\sim$(w/o $T$)                               & 10.01                           & 20.93                           \\
\gpt                                               & 9.95                            & 19.64                           \\
\gptt                                              & \textbf{16.23} & \textbf{29.65} \\ \hline
\multicolumn{1}{c}{\quarel-\cf}     &                                 &                                 \\ \hline
\textsc{lstm} Seq-to-Seq                         & 13.05                           & 24.76                           \\
\gpt$\sim$(w/o $T$)                                & 20.20                           & 36.64                           \\
\gptt$\sim$(w/o $T$)                               & 26.98                           & 41.14                           \\
\gpt                                               & 25.48                           & 42.87                           \\
\gptt                                              & \textbf{35.20} & \textbf{50.57} \\ \hline
\multicolumn{1}{c}{\defeasible-\cf} &                                 &                                 \\ \hline
\textsc{lstm} Seq-to-Seq                         & 7.84                            & 17.50                           \\
\gpt$\sim$(w/o $T$)                                & 9.91                            & 20.63                           \\
\gptt$\sim$(w/o $T$)                               & 9.17                            & 9.43                            \\
\gpt                                               & 10.49                           & \textbf{21.79} \\
\gptt                                              & \textbf{10.52} & 21.19                           \\ \hline
\end{tabular}
\caption{Generation results for \ours with baselines for language model $\mathcal{M}$.
We find that context is essential for performance (w/o $T$).
We provide these baseline scores as a reference for future research.
}
\label{tab:gen-quality}
\end{table}

\subsection{Baseline Language Models}
\label{sec:baselines}

To reiterate, \ours is composed of (i) task formulation component and (ii) graph construction component, that uses a language model \lm to construct the \st-graph. 
We want to emphasize that any language model architecture can be a candidate for \lm.
Since our \cf-task formulation is novel, we establish strong baselines for the choice of language model. 
Our experiments include large-scale language models (LSTM and pretrained transformer) with varying parameter size and pre-training, along with corresponding ablation studies. 
Our \lm choices are as follows: 

\vspace{0.5em}
\noindent \textbf{LSTM Seq-to-Seq:} 

We train an \textsc{lstm}~\cite{hochreiter1997long} based sequence to sequence model~\cite{bahdanau2015neural} which uses global attention described in~\cite{luong2015effective}.
We initialize the embedding layer with pre-trained 300 dimensional Glove~\cite{pennington2014glove}\footnote{https://github.com/OpenNMT/OpenNMT-py}. We use 2 layers of LSTM encoder and decoder with a hidden size of 500. The encoder is bidirectional.

\vspace{0.5em}
\noindent \textbf{GPT:} 
We use the original design of \gpt~\cite{radford2018improving} with 12 layers, 768-dimensional hidden
states, and 12 attention heads.

\vspace{0.5em}
\noindent \textbf{GPT-2:}
We use the medium (355M) variant of \textsc{gpt-2}~\cite{radford2019language} with 24 layers, 1024 hidden size, 16 attention heads.

For both \gpt and \gptt, we initialize the model with the pre-trained weights and use the implementation provided by~\citet{wolf2019huggingface}.

\subsection{Automated Evaluation}

To evaluate our generated \cf-graphs, we compare them with the gold-standard reference graphs.

To compare the two graphs, we first flatten both the reference graph and the \cf-graph as text sequences and then compute the overlap between them. 
We use the standard evaluation metrics \textsc{bleu}~\cite{papineni2002bleu}, and \textsc{rouge}~\cite{lin2004rouge}
\footnote{We use~\citet{sharma2017nlgeval} for calculating the overlap. \url{https://github.com/Maluuba/nlg-eval}}. 

Our results indicate that the task of \cf generation is challenging, and suggests that incorporating \cf-reasoning specific inductive biases might be beneficial.
At the same time, Table~\ref{tab:gen-quality} shows that even strong models like \gptt struggle on the \st-graph generation task, leaving a lot of room for model improvements in the future.

We also show ablation results for the model with respect to the context $T$~(\secref{subsec:task}), by fine-tuning without the context. We find that context is essential for performance for both \gpt and \gptt (indicated with w/o $T$ in Table~\ref{tab:gen-quality}). 

Further, we note that the gains achieved by adding context are higher for \gptt, hinting that larger models can more effectively utilize the context.

\subsection{Human Evaluation}
\begin{table}[!ht]
\setlength{\tabcolsep}{0.10em}
\centering
\begin{tabular}{lccc}
\toprule
Task & \gptt & \gptt & No \\
 &  (w/o $T$) &  &  Preference  \\
\midrule
Relevance & 23.05 & 46.11 & 30.83 \\
Reference & 11.67 & 31.94 & 56.39\\
\bottomrule

\end{tabular}
\caption{Results of human evaluation. The numbers show the percentage(\%) of times a particular option was selected for each metric.}
\label{tab:human_eval_relevance}
\end{table}

In addition to automated evaluation, we perform human evaluation on the ablation (\gptt- w/o $T$ and \gptt models) to assess the quality of generations, and the importance of grounding generations in context.
Three human judges annotated 120 unique samples for \textit{relevance} and \textit{reference}, described next. Both models (with and without context) produced grammatically fluent outputs without any noticeable differences.

\begin{table*}[!ht]
\centering
\small
\resizebox{\textwidth}{!}{%
\begin{tabular}{@{}llllll@{}}
\toprule
 Error Class & Description & \% & Question & Reference & Predicted \\ \midrule
 Polarity & The predicted polarity was wrong  & 5\% & What does `oil fields over-used' & there is not  & more oil \\
 & but event was correct &  & help at eventually ? & oil refined & is refined \\ \midrule
 Linguistic & The output was a & 20\% & What does `fewer rabbits will & more  & more \\
 Variability & linguistic variant of the reference &  & become pregnant' hurts at imminently ? &  rabbits & babies \\ \midrule
 Related  & The output was related but  & 17\% &  What does you inhale more air  & there will be & you develop  \\
 Event & different reference expected &  & from the outside hurts at imminently ? & less oxygen  & more blood clo-  \\ 
 & & & &in your blood& -ts in your veins \\ \midrule
 Wrong & The output was  & 30\% & What does `less nutrients for & more  & more wine  \\
 & was completely unrelated &  & plants' hurt at eventually ? & plants & being produced \\ \midrule
 Erroneous  & The gold annotations & 2\% & What does `less rabbit & less  & more  \\
 Reference & were erroneous &  & rabbit mating' hurt at imminently? & rabbits & babies \\ 
\bottomrule
\end{tabular}%
}
\caption{Examples of error categories. Error analysis is only shown for the incorrect outputs.}
\vspace{-0.5em}
\label{tab:gen-error-analysis}
\end{table*}

\paragraph{Relevance:} 

The annotators are provided with the 
input of a procedural text $T$, the \cf, and the relational questions.
The output events generated by \gptt (w/o $T$) and \gptt are also provided in random order. 
The annotators were asked, ``Which system (A or B) is more accurate relative to the background information given in the context?''
They could also pick option C (no preference).

\paragraph{Comparison with true event (reference):} We measure how accurately each system-generated event reflects the reference (true) event. 
Here, the annotators saw only the reference sentence and the outputs of two systems (A and B) in a randomized order. 
We asked the annotators, ``Which system's output is closest in meaning to the reference?'' 
The annotators could pick the options A, B, or C (no preference). 

For relevance and reference comparison tasks (Table \ref{tab:human_eval_relevance}), we present the percentage of the count of human judges for each of the three categories.
The table illustrates that \gptt performs better than \gptt (w/o $T$) on both the metrics.
Particularly, \gptt not only performs better than \gptt (w/o $T$) but also much better than the ``No Preference'' option in the relevance metric.
This means that \gptt generates target events that logically follow the passage and source events.
The reference and relevance task scores together show that \gptt does not generate target events that are exactly similar to the reference target events, but they are correct in the context of the passage and the source event.
This can happen due to linguistic variation in the generation, as well as the ability of the source event to influence multiple target events in the context of the passage.
We study this in more detail in the error analysis presented below.

\subsection{Error Analysis}
\label{subsec:error-analysis}
%\vspace{-0.5em}
Table~\ref{tab:gen-error-analysis} shows the error analysis on 100 random samples from the validation set.
We found that for about 26\% of samples, the generated event influence had an exact match with the reference, and about 30\% of the samples had no overlap with the reference (category \emph{Wrong} in Table \ref{tab:gen-error-analysis}). 
We found that for 20\% of the cases, the generated target event was correct but was expressed differently compared to the reference text (\emph{Linguistic Variability} class in Table \ref{tab:gen-error-analysis}).
Furthermore, we observed that in 17\% of cases, the generated target event was not the same as the reference target event, but was relevant to the passage and the question, as shown in the \emph{Related Event} category in Table~\ref{tab:gen-error-analysis}.
In 5\% of the samples (\emph{Polarity}), the model generates events with opposite polarity compared to the reference. 
A small fraction (2\%) of samples had incorrect gold annotations.

\subsection{Consistency Analysis}
\label{subsec:consistency}
Finally, we measure if the generated \cf-graphs are consistent. Consider a path of length two in the generated \cf-graph (say, A \green{$\rightarrow$} B \green{$\rightarrow$} C). A consistent graph would have identical answers to \textit{what does A help eventually} i.e., ``C'', and \textit{what does B help imminently} i.e., ``C''. 
To analyze consistency, we manually evaluated 50 random generated length-two paths, selected from \wiqa-\cf development set. We observed that 58\% of the samples had consistent output w.r.t to the generated output. We also measure consistency w.r.t. the gold standard, and observe that the system output is about 48\% consistent. Despite being trained on independent samples, our \cf-graphs show reasonable consistency and improving consistency further is an interesting future research direction.

\subsection{Discussion}
In summary, our task formulation allows adapting pretrained language models for generating \st-graphs that humans find meaningful and relevant.
Automated metrics show the utility of using large-scale models and grounding the \st-graph generation in context.
We establish multiple baselines with varying levels of parameter size and pretraining to guide future research.
\section{\rqtwo: \ours for Downstream Tasks}
\label{sec:downstreamqa}

In this section, we describe the approach for augmenting \cf graphs for downstream reasoning tasks.
We first identify the choice of tasks (\st-tasks) for domain adaptive pretraining~\citep{gururangan2020don} and obtain \ours language model $\mathcal{M}$.  
The downstream task then provides input context, \cf and (relation, type) tuples of interest, and obtains the \cf-graphs~(see Algorithm~\ref{alg:alg-1}). We describe one such instantiation in the section \secref{subsec:wiqa-qa}.

\subsection{\ours augmented \wiqa-\qa}
\label{subsec:wiqa-qa}

We examine the utility of \ours-generated graphs in the \wiqa-\qa~\cite{tandon2019wiqa} downstream question answering benchmark. Input to this task is a context supplied in form of a passage $T$, a starting event $c$, an ending event $e$, and the output is a label \{\textit{helps}, \textit{hurts}, or  \textit{no\_effect}\} depicting how the ending $e$ is influenced by the event $c$.

We hypothesize that \ours can augment $c$ and $e$ with their influences, giving a more comprehensive picture of the scenario compared to the context alone.
We use \ours trained on \wiqa-\cf to augment the event influences in each sample in the \qa task as additional context.

More concretely, we obtain the influence graphs for $c$ and $e$ by defining $R_{fwd} =$ \{(\textit{helps}, \textit{imminent}), (\textit{hurts}, \textit{imminent}) \} and $R_{rev} =$ \{ (\textit{helped by}, \textit{imminent}), (\textit{hurt by}, \textit{imminent})\}, and using algorithm~\ref{alg:alg-1} as follows: 
\begin{align*}
    G(c) &= \texttt{IGEN}(T, c, R_{fwd}) \nonumber \\ 
    G(e) &= \texttt{IGEN}(T, e,R_{rev})
\end{align*}

We hypothesize that \wiqa-\cf graphs are able to generate reasoning chains that connect $c$ to $e$, even if $e$ is not an immediate consequence of $c$. 
%\paragraph{Model:} 
Following \citet{tandon2019wiqa}, we encode the input sequence $\texttt{concat}(T, c, e)$ using the \textsc{bert} encoder $E$~\cite{devlin-etal-2019-bert}, and use the \textsc{[cls]} token representation ($\mathbf{\hat{h}}_i$) as our sequence representation.

We then use the same encoder $E$ to encode the generated effects $\texttt{concat}(G(c), G(e))$, and use the \textsc{[cls]} token to get a representation for augmented $c$ and $e$ ($\mathbf{\hat{h}}_a$).
Following the encoded inputs, we compute the final loss as: $\mathbf{l}_i = \texttt{MLP\textsubscript{1}}(\mathbf{\hat{h}}_i)$, and  $\mathbf{l}_a = \texttt{MLP\textsubscript{1}}(\mathbf{\hat{h}}_a)$ and $\mathcal{L} = \alpha \times \mathcal{L}_i + \beta \times \mathcal{L}_a$,
where $\mathbf{l}_i, \mathbf{l}_a$ represent the logits from $\mathbf{\hat{h}}_i$ and $\mathbf{\hat{h}}_a$ respectively, and $\mathcal{L}_i$ and $\mathcal{L}_a$ are their corresponding cross-entropy losses. $\alpha$ and $\beta$ are hyperparameters that decide the contribution of the generated influence graphs and the procedural text to the loss. We set $\alpha = 1$ and $\beta = 0.9$ across experiments.

\begin{table}[t]
\centering

\begin{tabular}{ccc} 
\toprule
Query Type &       \bert + \ours         & \bert       \\ 
\midrule
1-hop      & \textbf{78.78}  & 71.60  \\
2-hop      & \textbf{63.49}          & 62.50  \\
3-hop      & \textbf{68.28}  & 59.50  \\
\midrule
Exogenous  & \textbf{64.04}  & 56.13        \\
In-para &     73.58 & \textbf{79.68}     \\
Out-of-para & \textbf{90.84} &89.38 \\ 
\midrule
Overall & \textbf{76.92} & 73.80 \\
\bottomrule      
\end{tabular}
\caption{QA accuracy by number of hops, and question type. \bert refers to the original \bert results reported in \citet{tandon2019wiqa}, and \bert + \ours are the results obtained by augmenting the QA dataset with the influences generated by \ours.}
\label{tab:accuracy-hops}
\end{table}

\paragraph{QA Evaluation Results}

Table~\ref{tab:accuracy-hops} shows the accuracy of our method vs. the vanilla \bert model by question type and number of hops between $cf$ and $e$.
We also observe from Table~\ref{tab:accuracy-hops} that augmenting the context with generated influences from \ours leads to considerable gains over \bert based model, with the largest improvement seen in 3-hop questions (questions where the $e$ and $c$ are at a distance of three reasoning hops in the influence graphs).
The strong performance on the 3-hop question supports our hypothesis that generated influences might be able to connect two event influences that are farther apart in the reasoning chain.
We also show in Table~\ref{tab:accuracy-hops} that augmenting with \ours improves performance on the difficult exogenous category of questions, which requires background knowledge.

In summary, the evaluation highlights the value of \ours as a framework for improving performance on downstream tasks that require counterfactual reasoning and serves as an evaluation of the ability of \ours to reason about \cf-scenarios.

\subsection{Zero-shot Evaluation}
\label{sec:zeroshot}

In addition to supervised augmentation, we also evaluate \ours-$\mathcal{M}$ in a zero-shot setting.
Towards this, we perform a zero-shot evaluation on \quartz~\cite{Tafjord2019QuaRTzAO}, a dataset for qualitative counterfactual reasoning.
Each sample in \quartz consists of a question $\V{q}_i$ = \textit{If the top of the mountain gets hotter, the ice on the summit will:}, context $\V{k}_i$ = \textit{ice melts at higher temperatures}, the task is to pick the right answer from two options $\V{a}_i^1$ = \textit{increase}, and $\V{a}_i^2$ = \textit{decrease}. Since this task is setup as a qualitative binary classification task, \ours cannot be directly adopted to augment the QA pairs like described in Algorithm~\ref{alg:alg-1}.   

For the zero-shot setting, we use \ours-\lm fine-tuned on \quarel-\cf as our language model. 
For an unseen test sample $(\V{q}_i, \V{a}_i^1, \V{a}_i^2, \V{k}_i)$,  we  select $\V{a}_i^1$ as the correct answer if $p_{\theta}(\V{a}_i^1 \given \mathbf{x}_i) > p_{\theta}(\V{a}_i^2 \given \mathbf{x}_i)$, and select $\V{a}_i^2$ otherwise (here $p_{\theta}$ stands for \quarel-\cf). 
Our zero-shot \ours-$\mathcal{M}$ achieves a 54\% accuracy compared to supervised BERT model which achieves 54.7\% accuracy. 
These results suggest that \ours performs competitively at tasks while having no access to any supervision.

\subsection{Discussion}
In summary, we show substantial gains when a generated st-graph is fed as an additional input to the \qa model.
Our approach forces the model to reason about influences within a context, and then ask questions, which proves to be better than asking the questions directly.

\section{Related Work}
\label{sec:rel_wrk}

% \paragraph{Counterfactual Reasoning :} 
% A commonly studied situational reasoning paradigm is to predict the outcome in a context under a counterfactual situation. 
% Counterfactual reasoning is the ability to imagine the consequences of something that is contrary to what actually happened, thus diverging from the observed narrative. \cf reasoning has been studied across diverse disciplines including human psychology \cite{psychology-Epstude2008}, logic \cite{Lewis1979CounterfactualDA} among other disciplines. In the NLP community, there is a growing body of work on \cf reasoning \citep{nlp-counterfactual-Hobbs2005, Weld1990ReadingsIQ, tandon2019wiqa, counterfactual-story-generation-2019-emnlp,Fajcik2020BUTFITSemeval-detecting-counterfactuals}. Counterfactual reasoning has also been studied in the context of causal inference, \citet{pearl2000models} discusses causal interventions in the latent chain of events of the story. He distinguishes causal intervention from correlation by defining a $do(X)$ operator. Correlation is $P(X|Y)$ whereas causality is $P(X|do(A))$  where $A$ is an intervention. 
% The other line of work in NLP has been on what-if perturbations and their effect, under which we will now discuss two broad categories: closed domain and open domain \cf reasoning. 

\paragraph{Closed-domain \cf reasoning :}
In NLP, a large body of work has focused on what-if questions where the input is a context, \cf, and an ending, and the task is to predict the reachability from \cf to the ending. The most common approach \citep{tandon2019wiqa,Rajagopal2020WhatifIA,tafjord2019quarel} is a classification setting where the path is defined as more or less (qualitative intensities) over the sentences in the input context (a paragraph or procedural text with ordered steps). Such models do not generalize across domains because it is difficult to deal with changing vocabularies across domains. In contrast, our framework combines such diverse \cf-reasoning tasks under a general framework.

\paragraph{Open-domain \cf reasoning :}
Very recently, there has been interest in \cf reasoning from a retrieval setting \citep{Lin2019ReasoningOP} and a more common generation setting, attributed partially to the rise of neural generation models \citep{generation-models}. \citet{counterfactual-story-generation-2019-emnlp} presents generation models to generate the path from a counterfactual to an ending in a story. Another recent dataset \cite{rudinger-etal-2020-thinking} proposes defeasible inference in which an inference (X is a bird, therefore X flies) may be weakened or overturned in light of new evidence (X is a penguin), and their dataset and task is to distinguish and generate two types of new evidence -- intensifiers and attenuators. We make use of this dataset by reformulating their abductive reasoning setup into a deductive setup (see \S{\ref{subsec:dataset}} for details). 

Current systems make some simplifying assumptions, e.g. that the ending is known. Multiple \cf (e.g., more sunlight, more pollution) can happen at the same time, and these systems can only handle one situation at a time. Finally, all of these systems assume that the \cf happens once in a context. Our framework strengthens this line of work by dropping that assumption of an ending being given, during deductive \cf reasoning. In principle, our formulation is general enough to allow for multiple \cf and recursive reasoning as more situations unfold. Most importantly, our framework is the first to allow for \cf reasoning across diverse datasets, within a realistic setting where only the context and \cf are known.

\section{Conclusion}
\label{sec:conclusion}

We present \ours, a situational reasoning that: (i) is effective at generating \cf-reasoning graphs, validated by automated metrics and human evaluations, (ii) improves performance on two downstream tasks by simply augmenting their input with the generated \st graphs.
Further, our framework supports recursively querying for any node in the \cf-graph.
For future work, we aim to design advanced models that seeks consistency, and another line of research to study recursive \cf-reasoning as a bridge between dialog and reasoning.

% \newpage
\bibliography{acl2021}
\bibliographystyle{acl_natbib}
\end{document}